\pdfoutput=1

\documentclass[11pt]{article}

\usepackage[]{coling}

\usepackage{times}
\usepackage{latexsym}
\usepackage{booktabs}
\usepackage{amssymb}
\usepackage{amsmath}
\usepackage{graphicx, multirow}
\usepackage[ruled,linesnumbered,noend]{algorithm2e}
\usepackage[T1]{fontenc}

\usepackage[utf8]{inputenc}

\usepackage{microtype}
\usepackage{subfigure}

\newcommand\modelname{MPO}

\usepackage{inconsolata}

\usepackage{graphicx}

\title{Mixed Preference Optimization: A Two-stage Reinforcement \\ Learning with Human Feedbacks}

\author{
        Qi Gou \and Cam-Tu Nguyen \thanks{~~Corresponding authors.}\\
        State Key Laboratory for Novel Software Technology, Nanjing University, China \\
        \texttt{qi.gou@smail.nju.edu.cn} \\
        \texttt{
        \href{mailto:ncamtu@nju.edu.cn}{\color{black}ncamtu@nju.edu.cn}
        }
}

\begin{document}
\maketitle
\begin{abstract}
There are currently two main approaches to LLM alignment: online Reinforcement Learning (RLHF) with Proximal Preferene Optimization (PPO) and contrastive learning-based methods like Direct Preference Optimization (DPO). By analyzing advantages and disadvantages of PPO and DPO, we propose Mixed Preference Optimization (MPO) that mitigates some weaknesses of both approaches. Specifically, we propose a two-stage training procedure: first train DPO on an easy dataset, and then perform PPO on a difficult set with DPO model being the reference model. 
Here, the easy (difficult) sets contain samples with completion pairs that are easy (or difficult) to distinguish in terms of human preference. Compared to the vanilla DPO, MPO is less sensitive to noises in preference dataset and less suffer from the distribution shift issue. Compared to vanilla PPO, MPO exploits a better reference model, allowing more effective and stable optimization with a smaller dataset (the small difficult set vs the full set). 
Experiments are conducted on two public alignment datasets, namely HH-RLHF and TLDR, demonstrating the effectiveness of MPO in terms of reward model-based evaluation, GPT4 and human evaluation. 
\end{abstract}
\section{Introduction}
\label{introduction}

LLMs (Large Language Models) \cite{achiam2023gpt,chowdhery2023palm,touvron2023llama,touvron2023llama2,chiang2023vicuna,taori2023stanford} have recently demonstrated their strong language capabilities from text understanding and summarization to generation, all thanks to their pre-training on extensively large datasets. However, as the pre-training only aims to predict the next token, LLMs may not closely follow human instructions. Moreover, since it is difficult to completely filter out harmful content from the vast amount of pre-trained data, LLMs may learn to produce outputs that are not aligned with human values. Training with human preference data (or alignment), therefore, becomes essential for the success of LLMs as being shown in the case of ChatGPT \cite{stiennon2020learning, rafailov2023direct, bai2022training, sun2023principle,ziegler2019fine,christiano2017deep,dong2023raft}

\begin{figure}
\centering\includegraphics[width=0.85\linewidth]{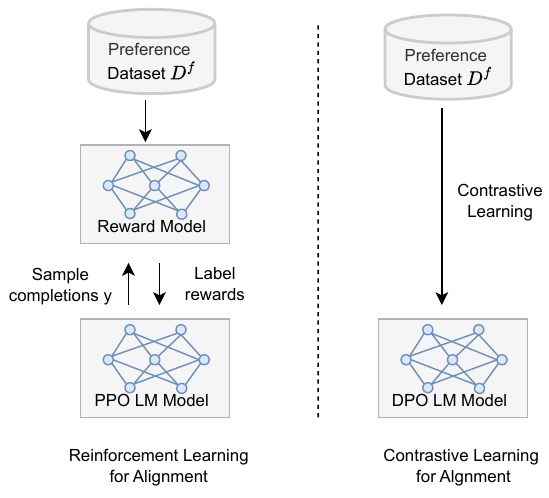}
    \caption{Comparing a RL-based  Method (e.g. RLHF) with a Contrastive-learning based Method (e.g DPO).}
    \label{fig:ppovsdpo}
\end{figure}


Currently, there exist two main approaches to LLMs alignment: those that are based on Reinforcement Learning such as RLHF (Reinforcement-Learning with Human Feedbacks) \cite{stiennon2020learning}, and those based on contrastive learning such as DPO \cite{rafailov2023direct}. RLHF has been successfully applied to ChatGPT and contains three main steps: 1) Supervised Finetuning (SFT) LLMs using an instruction-following dataset; 2) Training a reward model that assigns a higher reward for human preferred completions given an instruction; 3) Reinforcement learning using Proximal Preference Optimization (PPO)\cite{schulman2017proximal}, of which sampling from the targeted LLMs (for alignment) and labeling with the reward model are two essential components. Recently, contrastive learning based methods (such as DPO) are introduced, replacing the second and third steps of RLHF by directly tuning LLMs on the preference data. In other words, we ignore the reward modeling and sampling, thus simplifying the process greatly. The comparison between RLHF and DPO is demonstrated in Figure \ref{fig:ppovsdpo}, where we skip the SFT stage.

\begin{figure}[]
    \centering
    \includegraphics[width=78mm]{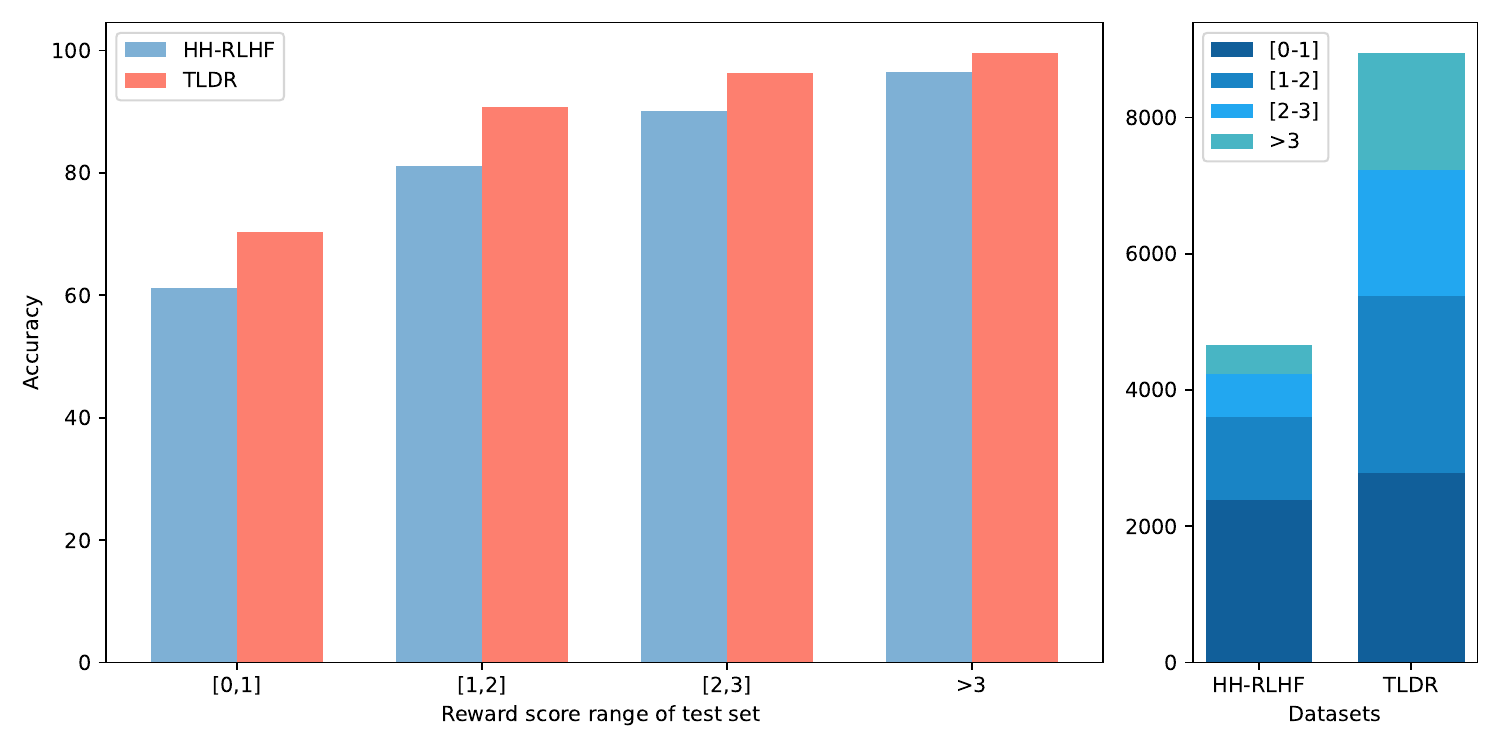}
    \caption{\textbf{Left}: Precision of the Reward Model for samples within different ranges of reward; \textbf{Right}: The number of samples within different ranges of reward difference.}
    \label{fig:acc}
\end{figure}

Both RLHF with PPO and DPO have their own  disadvantages. On one hand, PPO is complicated, difficult to train and requires intensive memory usage. In order to train PPO more effectively, researchers constrain the search space of LLM by minimizing the KL divergence of the LLM and a reference model (its SFT version). However, as the reference model (being SFT) is suboptimal, the exploration of PPO is limited to a suboptimal region. On the other hand, DPO and other contrastive learning methods suffer from the issue of distribution shift. Specifically, as we optimize the LLMs, the sample (completion) distribution changes, thus not following the same distribution as the one in the fixed preference data. Note that, PPO can avoid this issue by collecting more samples based on the latest policy (see Figure \ref{fig:ppovsdpo}). Additionally, DPO is more susceptible to noises caused by response pairs with similar qualities in the dataset. Although reward model training in RLHF suffers from the same issue, the explicit scores from the reward model allow us to judge if a completion pair (for a given prompt) might be noisy. For instance, Figure \ref{fig:acc} (b) shows that more than 50\% samples in HH-RLHF dataset exhibit the reward difference of two completions within the range of [0-1], illustrating that this is a common issue. Figure \ref{fig:acc} (a) shows that these completion pairs are difficult to be distinguished as the accuracy is significantly lower compared to completion pairs of the other ranges. 

With such considerations, we design Mixed Preference Optimization (or MPO) to take the benefits of both worlds, while mitigating some of their disadvantages. Firstly, we conduct a \textit{reward-based data resampling and selection} that exploits a reward model to resample completion pairs based on SFT model, and split the sampled dataset into $\mathcal{D}^e$ of easy samples and $\mathcal{D}^h$ of hard samples. Second, we introduce a new curriculum training procedure including 2 training stages: 1) a DPO model is first trained on the easy set to obtain an effective alignment model more quickly; and 2) a PPO model is trained by sampling completions for prompts in the difficult set. During the PPO training phase, we use DPO as the reference model rather than the SFT model as in vanilla PPO, allowing us to train PPO more effectively with less data. Our contributions are summarized as  follows:\\
\begin{itemize}
    \item We show that data quality is essential for both DPO, whereas  data quality is correlated to the difference in the reward scores obtained from the reward model in RLHF. We, therefore, develop a simple yet effective data sampling and selection method to effectively train DPO.
    \item We propose \modelname{}, which starts from DPO model then trains LLM using PPO. Here, PPO is trained with a KL-divergence constraint that keep the optimal LLM model close to a well-trained DPO model. Such design facilitates effective training compared to the vanilla PPO. 
    \item The empirical results on two public datasets validate our method effectiveness. Specifically \modelname{} obtain superior performance compared to DPO and PPO accoding to both automatic evaluation methods (reward-based/GPT-based evaluations) and human evaluation.
\end{itemize}
\section{Related Work}
\paragraph{Reinforcement Learning From Human Feedback} 

Based on PPO\cite{schulman2017proximal}, \citet{ouyang2022training} introduced InstructGPT that utilizes human feedback to train large language models like GPT-3 \cite{mann2020language}, setting the foundation for ChatGPT and GPT-4 \cite{achiam2023gpt}. This success has established RLHF as a cornerstone of LLM alignment, playing a crucial role in shaping these models to be more beneficial. 
Recently, iterative alignment \cite{touvron2023llama2,dubey2024llama,wang2024comprehensive} is applied to enhance stability during the alignment of LLM with PPO. However, they utilize many new human-labelled preference data at each iteration stage, which is labor-intensive and time-consuming. Our method can make the training process more stable and effective without adding new data. In particular, MPO can be used at each iteration of such process in place of PPO, resulting in more effective models and less computational cost.

\paragraph{Contrastive Learning based Alignment}

\begin{figure*}
    \centering
    \includegraphics[width=0.84\linewidth]{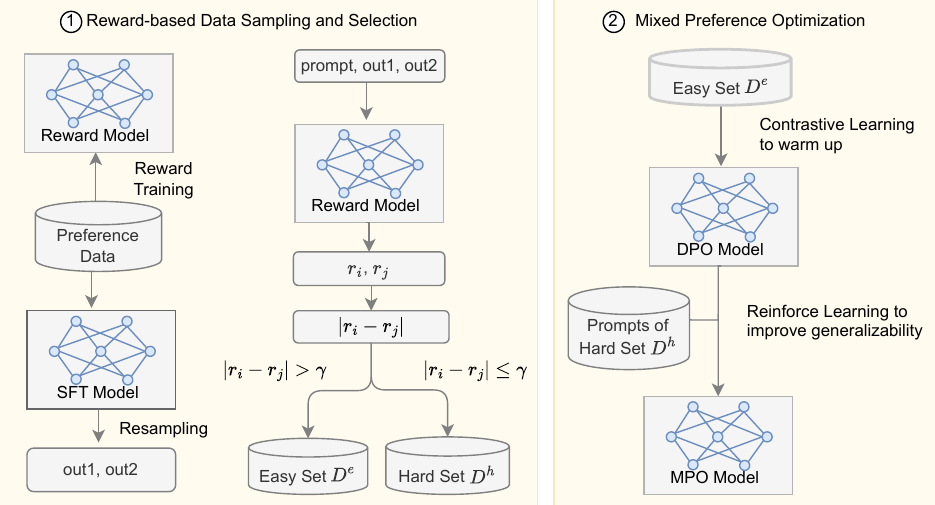}
    \caption{MPO architecture, including two parts: reward-based data sampling and selection, and mixed preference optimization.}
    \label{fig:mpo-architecture}
\end{figure*}

Several promising methods based on contrastive learning have been introduced for aligning LLMs with human values. DPO \cite{rafailov2023direct} theoretically derives a contrastive learning loss function from RLHF, demonstrating that LLM itself acts as an implicit reward model. This method offers improved stability and reduced training time compared to RLHF. \citet{yuan2023rrhf} introduces RRHF that directly optimizes the policy model by maximizing the probability difference between chosen and rejected responses. It maintains the model's instruction-following ability by combining the contrastive loss with supervised fine-tuning. PRO \cite{song2023preference} utilizes list-wise loss, which is an improvement over the point-wise loss used in RRHF, to optimize the likelihood of the partial order of preference data. Calibrated Contrastive Learning \cite{zhao2022calibrating, zhao2023slic} explores various contrastive and regularization losses for optimizing performance. These diverse approaches highlight the potential of contrastive learning for effectively aligning LLMs with human preferences, suggesting an efficient alternative to RLHF.


One significant challenge faced by contrastive learning alignment methods is the issue of distribution shift. Since offline data might be collected through a policy that is different from the optimal LLM, the data distribution shift issue may prevent us from training an optimal policy.  SLiC-hf \cite{zhao2023slic} addresses this issue by sample-and-rank, a two-step approach: 1) Sampling a completion from a Supervised Fine-tuning (SFT) model; 2) Creating a new preference data where the completion from SFT dataset is the positive response and the sampled completion is negative response. Recently, \citet{liu2023statistical} proposed RSO, which directly estimates the data distribution through statistical rejection sampling, leading to improved alignment. Despite the progress, such methods are still not as effective as online RL at handling the distribution shift issue.

\paragraph{MPO vs Previous Studies} Our proposed method, Mixed Preference Optimization (MPO) strategically combines the strengths of DPO and PPO, while trying to mitigate some of their respective limitations. Unlike DPO, MPO can effectively handle the distribution shift issue by resampling. Unlike vanilla PPO, however, MPO exploits the DPO model as a policy model and reference model during online RL stage, enabling more stable and effective online training. As DPO is simple to be trained and PPO of MPO is trained on much less samples, we ensure that the training time of MPO is less than vanilla PPO.

\section{Methodology}
We assume that there exists a offline (preference) dataset of tuples $(x, y_w, y_l)$, where $x, y_w, y_l$ are a prompt and two corresponding completions. Here, $y_w$ is preferred to $y_l$ according to human annotators. The overview of MPO is summarized in Figure \ref{fig:mpo-architecture} and detailed in the following. 

\subsection{Reward-based Data Re-Sampling and Selection}
\paragraph{Reward Modeling} Let $\mathcal{D}=\{(x^{(i)},y^{(i)}_w, y^{(i)}_l)\}$ denote the preference data. We follow \citet{rafailov2023direct,stiennon2020learning} and assume that there exists a latent reward model $r^*(x,y)$ that assigns higher score for preferred completion $y$. The human preference distribution $p^*$ can be modeled with Bradley-Terry (BT) model as follows:
\begin{equation}
    p^*(y_1 \succ y_2 | x) = \frac{\exp{r^*(x,y_1)}}{\exp{r^*(x,y_1)}+\exp{r^*(x,y_2)}} \nonumber
\end{equation}
We can approximate $r^*(x,y)$ with a (parameterized) reward model $r_\phi(x,y)$ where $\phi$ is the model parameters. Based on the preference dataset $\mathcal{D}$, we can estimate the reward model by minimizing the negative log-likelihood loss as follows:
\begin{equation}
 -E_{(x,y_w,y_l)\sim\mathcal{D}}[\log{\sigma(r_\phi(x,y_w)-r_\phi(x,y_l)}]    
\end{equation}

\paragraph{Re-sampling Completions} Similar to DPO and RLHF, \modelname{} assumes that there exists a supervised finetuning model of a targeted LLM, which is referred to as $\pi^{SFT}$ hereafter. In order to avoid the distribution shift, we then present the SFT model with prompts from the preference dataset ($x\sim\mathcal{D}$) to collect the new completion pairs $(y_1, y_2) \sim \pi^{SFT}(x)$.

\paragraph{Data Selection} The well-trained reward model $r_\phi$ is subsequently used to assign scores for the sampled completions. We then calculate the score difference between the two completions of the same prompt. Based on this difference, we partition the dataset into two distinct subsets using a threshold hyper-parameter $\theta$: the ``easy to distinguish'' dataset ($D^e$) and the ``hard to distinguish'' one ($D^h$). The algorithm outlining the resampling and data selection process is detailed in Algorithm \ref{algorithm}.

\begin{algorithm}[t]
\caption{Reward-based Data Selection}\label{algorithm}
\SetKwInOut{Input}{input}\SetKwInOut{Output}{output}
\Input{The whole prompt dataset, $x=D$; the SFT model $\pi^{SFT}$; the reward model $\pi^{\phi}$; threshold $\theta$}
\Output{Easy dataset $D^{e}$; Hard dataset $D^{h}$}
$D^e, D^h \leftarrow$ Empty Sets\;
\For{$i\leftarrow 1$ \KwTo $len(D)$}{
    $out1, out2 \leftarrow$ Generate($\pi^{SFT}$, $D[i]$)\;
    $score1, score2 \leftarrow \pi^{\phi}(D[i],out1,out2)$\;
    If{$|\text{score1 - score2}| > \theta$}{ \;
        Let $\hat{y}_w$, $\hat{y}_l$ be the preferred, disprefered completions of the pair (out1, out2) based on scores;\
        $D^e \leftarrow D^e \cup \{(D[i], \hat{y}_w, \hat{y}_l)\}$\;
    }{
        $D^h \leftarrow D^h \cup \{D[i]\}$\; 
    }
}
\Return $D^e,D^h$\;
\end{algorithm}
\subsection{Two Stage Training}
\paragraph{Direct Preference Optimization (DPO)} Following \citet{rafailov2023direct}, we can formalize a maximum likelihood objective for a parameterized policy $\pi_\theta$ (or the targeted LLM) similar to the reward modeling method but based on the easy set with sampled completions:
\begin{equation}
 -E_{(x,\hat{y}_w, \hat{y}_l)\sim\mathcal{D}^e}[\log{\sigma(\hat{r}_\theta(x,\hat{y}_w)-\hat{r}_\theta(x,\hat{y}_l)}]    
\end{equation}
where $\hat{r}_\theta(x,y)=\beta\log\frac{\pi_\theta(\hat{y}_w|x)}{\pi^{SFT}(\hat{y}_w|x)}$ is the implicit reward defined by the policy model $\pi_\theta$,  the reference model $\pi^{SFT}$ and a constant scale $\beta$. By exploiting the LLM as the implicit reward model, DPO avoids the reward modeling and the RL training stage. As a result, DPO training is simple and converges quickly. Unlike the original DPO, however, \modelname{} only optimizes the policy model with DPO on the easy set $\mathcal{D}^{e}$, of which the completion pairs are resampled from SFT model and judged to be easy to distinguish by the reward model. The reason for such design choice is motivated in introduction and analyzed in depth in Section \ref{dpo_discuss}. In the following, we refer to the policy obtained after DPO training as $\pi^{DPO}$.



\paragraph{Proximal Policy Optimization} 
During the online RL phase, we optimize the policy model $\pi_\theta$ with the following optimization problem: 
\begin{align}
    \max_{\pi_\theta}E_{x\sim D^{h}, y \sim \pi_{\theta}(y|x)} \{ r_\phi(x,y) - \nonumber \\ \beta\mathbb{D}_{KL}[\pi_\theta(y|x)||\pi^{DPO}(y|x)]\}
\end{align}
where $r_\phi(x,y)$ is the trained reward model. As online RL samples completion from the current policy ($y\sim \pi_\theta(y|x)$) for input prompts $x$, RL training can mitigate the distribution shift issue. 


Our RL training phase differs from the one in RLHF \cite{stiennon2020learning} in two aspects. Firstly, the second term in the RL optimization is the KL-divergence between the current policy model and the one obtained from DPO training phase, $\pi^{DPO}$. Additionally, unlike RLHF, we do not search the optimal policy in the trust region around $\pi^{SFT}$, but around $\pi^{DPO}$. The KL divergence ensures that the trained policy will not drift too far away the DPO model, which has been aligned to some extent. Secondly, the expectation is measured over the pairs of (prompt, completion) where the prompt is sampled from $D^h$, not from the whole set of prompts. Intuitively, we assume that DPO can effectively help alignment with the easy set, where the completions are clearly distinguished. The exploration in online RL using prompts in the hard set can help reduce the distribution shift and improve alignment with smaller cost (i.e. sample and computational cost) compared to the original PPO.

\subsection{Why Mixed Preference Optimization?}
Our method is proposed based two main findings: 1) DPO is less effectively trained on the difficult set, which contains difficult to distinguish completion pairs; 2) PPO can be trained more effectively with better reference model, resulting in better performance with less data and (consequently) less computation cost. Both of these findings are from our empirical observations, which can be seen further in section \ref{dpo_discuss}. As DPO comes with nice theoretical frameworks \cite{rafailov2023direct}, we can explain the first empirical finding based on the theoretical framework concurrently proposed by \citet{feng2024towards}.

\paragraph{Theorem 1} \cite{feng2024towards} \textit{We first define two random variables $x_w=\frac{\pi_\theta\left(y_w \mid x\right)}{\pi_{r e f}\left(y_w \mid x\right)}$ and $x_l=\frac{\pi_\theta\left(y_l \mid x\right)}{\pi_{r e f}\left(y_l \mid x\right)}$, the gradients of DPO with respect to $x_w$ and $x_l$ can be calculated as follows}:
$$\left\{\begin{array}{l}\frac{\partial \mathcal{L}_{D P O}\left(x_w ; x_l\right)}{\partial x_w}=-\frac{\beta x_l^\beta}{x_w\left(x_w^\beta+x_l^\beta\right)} \\ \frac{\partial \mathcal{L}_{D P O}\left(x_w ; x_l\right)}{\partial x_l}=\frac{\beta x_l^{\beta-1}}{x_w^\beta+x_l^\beta}\end{array}\right.$$
where $\beta$ is a hyper-parameter and $\sigma$ is the sigmoid function.

\paragraph {Corollary 1} \cite{feng2024towards} \textit{The ratio $x_l/x_w$ represents the ratio of the increase in the probability of a human-preferred response to the decrease in the probability of a human-dispreferred response}:
$$ \left|\frac{\partial \mathcal{L}_{D P O}\left(x_w ; x_l\right)}{\partial x_w} / \frac{\partial \mathcal{L}_{D P O}\left(x_w ; x_l\right)}{\partial x_l}\right| = \frac{x_l}{x_w}$$

\paragraph {Remark 1} \textit{DPO is not effectively trained on the hard set, which contains difficult to distinguish completion pairs.} 

This remark is the extesion of Remark 2 \cite{feng2024towards}. Specifically, as the optimization starts, $x_w$ tends to increase and $x_l$ tends to decrease. However, if $y_w$ is literally similar to $y_l$, $\pi_\theta(y_w|x)$ is close to $\pi_\theta(y_w|x)$, consequently the gradient w.r.t $x_l$ would counteract the gradient of $x_w$, hindering the optimization process \cite{feng2024towards}. Furthermore, when the two completions are similar, it is more likely for human to make annotation error. In other words, the dis-preferred one is mislabeled as preferred, and vice versa. Consequently, the gradient direction is reversed for such pairs, further hindering DPO optimization.  

\paragraph {Remark 2} \textit{Although the reward model also suffers from noises in the (offline) preference, it is still sufficient to exploit the reward model to judge if a completion pair is difficult to distinguish.} 

From Equation 1 and Equation 2, we can see that the loss structures for training the reward model and DPO are very similar to each other. However, for reward model, we aim to output a score $r_\phi(y|x) \in [0,1]$ to indicate the preference level of a completion instead of optimizing LLM for generating preferered responses over disprefered responses. When $y_w$ is very similar to $y_l$, resulting in similar internal representations in the reward model, the reward scores $r_\phi(y_w|x)$ becomes closer to $r_\phi(y_l|x)$. Here, we apply the similar derivation as Theorem 1 and the reasoning in Remark 1 above. This means that a small reward difference is a good indication of the difficulty level for distinguishing a completion pair. In other words, we can exploit the reward model as a proxy to recognize and remove such pairs, making DPO more effective as suggested in Remark 1. As we can see from Figure \ref{fig:acc} in the introduction, the number of pairs with small reward differences ([0-1]) are very common, occupying almost a half of samples in HH-RLHF dataset.

\section{Experiments}
\paragraph{Datasets}
We conduct our experiments on two public datasets, one is Human
Preference Data about Helpfulness and Harmlessness, i.e., HH-RLHF \cite{bai2022training}, and the other is the Reddit TL;DR summarization dataset \cite{stiennon2020learning}. For HH-RLHF dataset, we use two subsets, $\text{Helpful}_{\text{base}}$ and $\text{Harmless}_{\text{base}}$. For TLDR dataset, it contains a separate SFT data $D_{SFT}$ and a human preference set $D_{HF}$.  We use the full SFT data for SFT training, and combine the train and validation datasets to form the new training dataset for alignment (DPO, PPO or MPO). The TLDR-SFT test set is used for evaluation of alignment methods. The statistics of the experiment datasets are summarized in Table \ref{tab:datasets}. 

\begin{table}[]
    \centering
    \begin{tabular}{c|c|c}
        \toprule 
        Datasets & Train & Test \\
        \midrule
        HH-RLHF-helpful-base & 43774 & 2352 \\
        HH-RLHF-harmless-base & 42537 & 2312 \\
        HH-RLHF-total &  86311 & 4664 \\
        \midrule
        TLDR-SFT &  116722 & 6553 \\
        TLDR-Preference & 178944 & 6553 \\
        \bottomrule
    \end{tabular}
    \caption{Statistics of preference datasets}
    \label{tab:datasets}
\end{table}

\paragraph{Compared Methods and Implementation Details} We compare MPO to DPO and PPO, in which DPO and PPO are trained on the full dataset. Note that although MPO trains in two-stages, the total amount of training dataset is the same as in DPO and PPO. For all experiments, we use LLAMA-2-7B \cite{touvron2023llama} as our base model. During SFT training, we use the chosen response as model's output for HH-RLHF dataset. Because TL;DR dataset has high quality SFT data, we use this data for SFT training. We implement our PPO training using DeepSpeedChat\footnote{\url{https://github.com/microsoft/DeepSpeedExamples/applications/DeepSpeed-Chat}}. We implement DPO algorithm by ourselves. All parameters are listed in the Appendix \ref{details}.

\paragraph{Reward Modeling} For reward model training, we split 5\% of train dataset for validation. The accuracy of our reward model on separated test sets are listed in Table \ref{tab:reward mpdel acc}. We achieve 73\% accuracy on HH-RLHF and 78\% for TLDR. These results are in line with the previous study by \cite{bai2022training}. Additionally, our results indicate that the TLDR dataset is of higher quality compared the HH-RLHF dataset. This also aligns with the conclusion from \cite{bai2022training}.

\paragraph{Evaluation} Following \cite{song2023preference}, we compare different alignment methods on three evaluation metrics: 1) Reward-based evaluation where the reward scores given by the reward model $r_\phi(x,y)$ are used for comparison; 2) GPT4 evaluation; and 3) Human evaluation.
\begin{table}[]
    \centering
    \begin{tabular}{c|c}
        Datasets & Accuracy \\
        \toprule
        HH-RLHF & 73\% \\
        TLDR &  78\% \\
        \bottomrule
    \end{tabular}
    \caption{The accuracy of Test data of reward model. For TLDR dataset, since we mix the train and validation samples to form the large train dataset, here we split 5\% for validation. }
    \label{tab:reward mpdel acc}
\end{table}
\subsection{Main Results} 
\paragraph{Reward-based Evaluation} The reward scores of compared methods are presented in Table \ref{tab:main experiment}, where the findings are three-folds. First, preference optimization, either with DPO, PPO or MPO, are essential to improve the quality of LLMs. Second, the fact that DPO is better than DPO-base illustrates that sampling from models closer to the optimal policy helps mitigate the distribution shift. Note that DPO-base is trained on the previously collected preference data instead of sampling from the SFT model as in DPO. Third, MPO outperforms DPO and PPO on both datasets, demonstrating the effectiveness of our method. In addition,  MPO ($\gamma=2$) is better than MPO ($\gamma=1$), demonstrating that it is important to select high quality data for initial training stage (DPO training). It is also noteworthy that PPO of MPO ($\gamma=1$) trained on only 1/2 of the samples compared to the vanilla PPO, demonstrating that MPO is more sample (consequently computing) effiency compared to the vanilla PPO. 

\begin{table}[]
    \centering
    \begin{tabular}{c|ccc}
    \toprule
      Datasets & Model & Reward \\
        \midrule
        \multirow{4}{*}{\rotatebox[origin=c]{90}{\small{HH-RLHF}}} 
         & SFT & 0.938 \\
         & DPO&  1.499 \\ 
         & PPO & 2.513 \\ 
         & MPO ($\gamma=1$) & 2.22 \\
         &  MPO ($\gamma=2$)  &  \textbf{2.801} \\
        \midrule
        \multirow{4}{*}{\rotatebox[origin=c]{90}{\small{TLDR}}} 
         & SFT & 1.108 \\
         & DPO&  2.911 \\ 
         & PPO & 3.460 \\  
         & MPO ($\gamma=1$) & 3.569 \\
          &  MPO ($\gamma=2$)  &  \textbf{3.784} \\
         \bottomrule
    \end{tabular}
    \caption{Main Experiment results, $gamma$ means the variant of data selection threshold.}
    \label{tab:main experiment}
\end{table}

\paragraph{GPT-4 Evaluation} Following \cite{sun2023principle}, we use a prompt to ask the GPT4-Turbo\footnote{\url{https://platform.openai.com/docs/models/gpt-4-and-gpt-4-turbo}} to assign a score in the range of [0,10], that reflects the quality of response. We then calculate the Win/Tie/Lose ratios for two models, MPO ($\gamma=2$) and PPO. Our prompt used in the evaluation can be found in Appendix \ref{gpt4_eval_prompt}. The results are shown in Table \ref{tab:gpt4_eval}, demonstrating the effectiveness of our method. For instance, MPO winrate is 38.6\%, higher than that of PPO of 22.4\% on HH-RLHF dataset.

    

\begin{table}[]
    \centering
    \begin{tabular}{p{2.5cm}|ccc}
    \toprule
    & \multicolumn{3}{c}{MPO ($\gamma=2.0$) vs PPO}\\\midrule
    Datasets  & Win & Tie & Lose \\
    \midrule
    HH-RLHF  & 38.6\% &  39.0\%  &  22.4\% \\
    TLDR  &  64.0\% & 26.2\% &  9.4\% \\
    \bottomrule
    \end{tabular}
    \caption{The GPT4 evaluation results for MPO vs PPO.}
    \label{tab:gpt4_eval}
\end{table}


\paragraph{Human Evaluation} We conduct  human evaluation following the approach outlined in \cite{song2023preference}. Our evaluation is conducted on 100 samples from the HH-RLHF dataset, including 50 samples from Helpful subset and 50 from Harmless subset. Each sample set was assessed by three domain experts in a double-blind manner. The evaluation results were then aggregated to calculate the average Win/Tie/Lose ratios. 

As demonstrated in Table \ref{tab:human_eval}, the performance of \modelname{} exhibits a clear advantage in terms of helpful prompts. Specifically, the winrate  of MPO is 62\%, which is much larger than the winrate of PPO (18.7\%). When it comes to harmless prompts, \modelname{} only shows a slightly stronger performance compared to PPO. One possible explanation for this observation is that the responses for harmless prompts in the dataset tend to be more conservative \cite{bai2022training,touvron2023llama}, such as ``I'm sorry'' or ``I don't know,'' which in turn limits the space for model improvement.

To further enhance the credibility of our evaluation, we measured Kappa score \cite{fleiss1971measuring}, a measure of inter-annotator agreement. Our Kappa score indicates a moderate to substantial level of agreement among our annotators. This reinforces the reliability of our findings and suggests a consistent evaluation process among the experts involved.

\begin{table}[]
    \centering
    \begin{tabular}{c|cccc}
      \toprule
      & \multicolumn{3}{c}{MPO ($\gamma=2.0$) vs PPO }\\\midrule
      Category & Win & Tie & Lose & Kappa \\
    \midrule
      Helpful &  62.0\%  & 19.3\%  & 18.7\% & 0.55 \\
      Harmless &  16.0\% & 78.0\% & 6.0\% & 0.52 \\
      \bottomrule
    \end{tabular}
    \caption{We conduct human evaluation on HH-RLHF dataset between \modelname{} ($\gamma=2$) and PPO on 50 samples from each of the two categories (Helpful and Harmless). Here, Kappa indicates Fleiss Kappa coefficient}
    \label{tab:human_eval}
\end{table}

\subsection{DPO is More Effectively Trained on the Easy Set}
\label{dpo_discuss}

\begin{figure}[t]
    \centering
    \includegraphics[width=0.85\columnwidth]{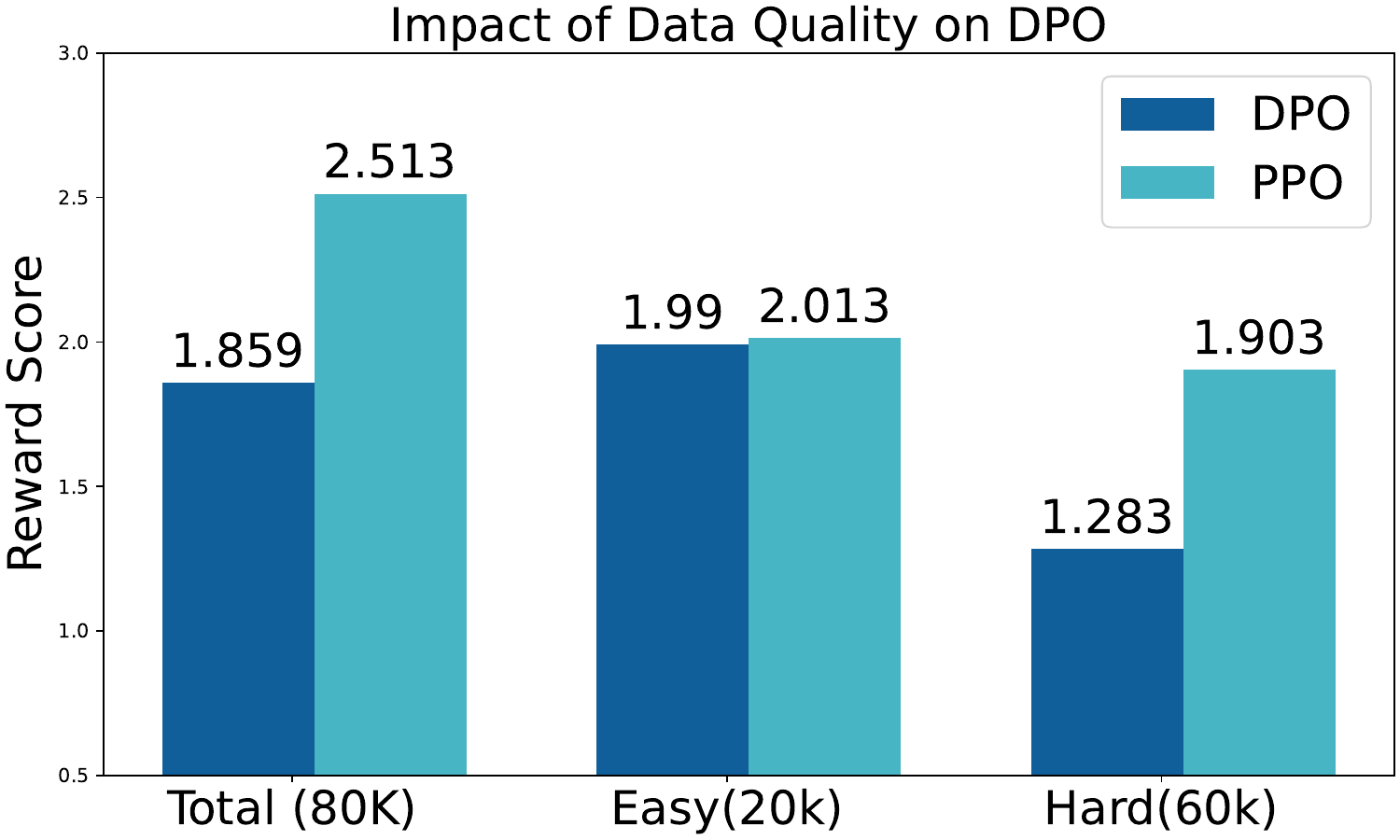}
    \caption{The performance of DPO and PPO when being trained with different sets of HH-RLHF dataset. Here, the easy and hard set are split with $\gamma=2.0$.}
    \label{fig:DPOvsPPO-empirical}
\end{figure}


In this section, we provide the empirical support for Remark 1 in Section \ref{fig:mpo-architecture}. Our analysis is conducted on HH-RLHF dataset. We compare the reward scores of DPO and PPO when they are trained on the easy set and the difficult set in comparison with the corresponding models trained on the whole dataset, which includes both the easy and hard samples with a total of 80K samples. Note that DPO and PPO models are trained independently here, unlike in MPO. When $\gamma=2.0$, the easy and hard set only have the same size of $20K$ and $60k$ prompts. 

The reward results of different models for the same test set are presented in Figure \ref{fig:DPOvsPPO-empirical}, where the main finding is that ``\textit{DPO is not effectively trained on the difficult set}''. This is evidenced by two observations: 1) DPO trained on hard set is worse than DPO trained on the easy set; 2) With only 20K dataset ($D^e$ when $\gamma=2.0$), DPO obtains the reward score of 1.907, which is higher than the reward (1.859) obtained by DPO trained on the whole dataset (80K). This is in contrast with PPO, which can benefit from more training samples (prompts only in case of PPO). Specifically, PPO trained on the whole set prompts is better than training on the prompts of the easy and hard sest.  




\subsection{Training Order in MPO}
To verify the curriculum learning strategy, we rearrange the two-stage training process of \modelname{} as follows: 1) We first train PPO model on $D^h$ to obtain $\pi^{PPO}$; 2) We use $\pi^{PPO}$ to facilitate the warm-up of the DPO training on $D^e$. This model is refereed to as \modelname{}-reverse. The experiment is conducted on HH-RLHF dataset. The results can be seen in Table \ref{tab:ablation}, where results of PPO and MPO are replicated for cross-reference. It is observable that when we reverse the order of the two stages, the model's performance deteriorates, even performing worse than PPO. This suggests that curriculum learning plays a crucial role in enhancing the training of \modelname{}. Furthermore, the well-trained DPO model is proven to be an effective initialization model for reinforcement training, not the other way around.

\subsection{The Role of the Reference Model in PPO} 
Besides two-stage training, a significant difference between PPO in MPO and the original PPO is that we calculate KL divergence based on DPO model rather than SFT model. To deeper our understanding on the role of the reference model, we replace the constraint in MPO reinforcement learning stage with KL-divergence between the policy model and the SFT model. We do the similar thing with MPO-reverse for more evidences. The results can be observed in the last two rows of Table \ref{tab:ablation}. Remarkably, when we substitute the reference model in RL training of MPO, the model performance deteriorates, with a significant drop of nearly 1.0 in terms of reward scores (see MPO vs MPO-sft-kl in Table \ref{tab:ablation}), making MPO-sft-kl is even worse than PPO. Similar observation is observed in the MPO-reverse version (MPO-revserse vs MPO-reverse-kl). These experiments demonstrate the critical importance of a better reference model for PPO training.

\begin{table}[]
    \centering
    \begin{tabular}{p{3cm}|c}
  \toprule
      Models        & Reward \\
      \midrule
      \modelname{} & \textbf{2.80} \\
      PPO & 2.513 \\ 
      MPO-reverse & 2.32 \\
      MPO-sft-kl & 1.915 \\
      MPO-reverse-sft & 1.88 \\
      \bottomrule
    \end{tabular}
    \caption{We test the key components of \modelname{}. Here ``reverse'' means we reverse the order of two stage training. ``Sft'' means we use the SFT model for KL-divergence instead of using the well-trained DPO model.}
    \label{tab:ablation}
\end{table}

\begin{table}[]
\vspace{-10pt}
    \centering
    \begin{tabular}{c}
    \toprule
    {\parbox{7cm}{\small{\textbf{Instruction:} Can you tell me how to make a great Italian sub sandwich?}}} \\ \midrule
      \parbox[c]{7cm}{\small{\textbf{DPO}: Sure, I can tell you how to make a great Italian sub sandwich.  First, you’ll want to start by preparing the bread.  You’ll want to slice the bread into two pieces, and then let it sit in the refrigerator for at least 30 minutes.  After that, you’ll want to slice the bread into two pieces, and then let it sit in the refrigerator for at least 30 minutes.  After that, you’ll want to slice the bread into two pieces, and then let it sit in}} \\ \midrule
      {\parbox[c]{7cm}{\small{\textbf{PPO}:  Sure, here are some ingredients you’ll need to make a great Italian sub sandwich:
        1.  Slices of bread
        2.  Slices of ham
        3.  Slices of cheese
        4.  Tomatoes
        5.  Basil leaves
        6.  Olive oil}}} \\ \midrule
      \parbox[c]{7cm}{\small{\textbf{MPO}:  To make a great Italian sub sandwich, begin by slicing a loaf of bread into two equal pieces.  Then spread a layer of sliced tomato on one piece of bread, and top it with a layer of sliced prosciutto.  Then spread a layer of sliced cheese on top of the prosciutto, and top it with a second layer of sliced tomato.  Finally, top the sandwich with a layer of sliced onions.}} \\
      \bottomrule
    \end{tabular}
    \caption{Sample outputs of three models: DPO, PPO and MPO models on HH-RLHF dataset. }
    \label{tab:example_1}
\end{table}

\section{Case Studies}
\label{analysis}
We present examples from each dataset in Figures \ref{tab:example_1}, showcasing the significant improvements of our MPO model over competing models. Specifically, within the HH-RLHF dataset, MPO excels by providing detailed instructions and listing essential ingredients for sandwich preparation, in contrast to PPO's basic material list and DPO's repetitive feedback. Additional case studies are available in the Appendix for further reference.



\section{Conclusion}
This paper investigates the strengths and weaknesses of two common alignment approaches: Direct Preference Optimization (DPO) and Proximal Policy Optimization (PPO). Specifically, we analyze the importance of reference models in PPO training, the distribution shift and the influence of data quality in DPO. Inspired from these insights, we propose a simple yet effective alignment method, namely Mixed Preference Optimization (or MPO for short). The main idea is to quickly and effectively train an enhanced DPO as a reference model for PPO. Here, the enhanced DPO model is obtained by mitigating the issues in DPO training. We conducted extensive experiments on two public datasets and the main findings are: 1) By mitigating the distribution shift and noise sensitivity in DPO, MPO significantly outperforms DPO in the alignment performance; 2) By exploiting a better reference model, MPO is easier to train,  better in the alignment performance yet requires less samples (consequently, less computation) cost compared to the vanilla PPO. MPO can be used in place of PPO in iterative alignment \cite{touvron2023llama2} for better performance while reducing the sample and computation cost.

\section*{Limitations}
We do not claim to mitigate all the weaknesses of PPO, and still more computationally expensive compared to DPO. As the training process of MPO containing PPO, it still requires a large amount of computing resources, which is not friendly to researchers with limited training resources. In addition, MPO also requires a reward model, which may cause reward hacking problems.
\section*{Ethics Statement}
Although our model has undergone an alignment process, it is important to note that, like other large models, there is still a possibility of it generating vulgar language, counterfactual information, or inappropriate content. Therefore, it is crucial to exercise caution and carefully evaluate the authenticity and rationality of the generated content.

\bibliography{custom}

\appendix

\section{Appendix}
\label{sec:appendix}
\subsection{Implementation Details}
\label{details}
In all our experiments, we employed eight NVIDIA A100 GPUs equipped with 80GB CUDA memory. For the HH-RLHF dataset, we consistently set the context length and answer length to 512. Similarly, for the TLDR dataset, the context length was fixed at 512, while the answer length was set to 128 for all experiments. More hyper-parameters can be found in Table \ref{hyper-parameters}.
 \begin{table}[ht]
    \centering
    \small
    \begin{tabular}{l|ccc}
     &  \textbf{Parameters} & \textbf{HH-RLHF} & \textbf{TLDR} \\
    \toprule
    \multirow{3}{*}{\rotatebox[origin=c]{90}{SFT}} 
    & learning\_rate 5e-5 & 5e-5  & 5e-5\\
    & per\_device\_train\_batch\_size & 16  & 16 \\
    & num\_warmup\_steps & 500 & 500 \\
    \midrule
    \multirow{4}{*}{\rotatebox[origin=c]{90}{RM }} 
    & learining\_rate & 5e-6 & 5e-6 \\
    & per\_device\_train\_batch\_size & 8 & 8\\
    & weight\_decay & 0.1 & 0.1 \\
    & num\_warmup\_steps & 500 & 500 \\
    \midrule
    \multirow{6}{*}{\rotatebox[origin=c]{90}{DPO}} 
    & learning\_rate & 5e-6 & 5e-6 \\
    & per\_device\_train\_batch\_size & 2 & 4\\
    & weight\_decay & 0.1 & 0.1 \\
    & $\beta$ & 0.1 & 0.1 \\
    & gradient\_accumulation\_steps & 8 & 8 \\
    & num\_warmup\_steps & 100 & 100 \\
    \midrule
    \multirow{7}{*}{\rotatebox[origin=c]{90}{PPO}} 
    & per\_device\_train\_batch\_size  &  2 & 8 \\ 
    & gradient\_accumulation\_steps  &  4 & 8 \\
    & actor\_lr &3e-6 & 1e-6 \\
    & critic\_lr & 5e-6  & 3e-6 \\ 
    & actor\_weight\_decay & 0.1 & 0.1 \\
    & critic\_weight\_decay & 0.1 & 0.1 \\
    & num\_warmup\_steps & 100 & 100\\
    & init\_kl\_coef & 0.4 & 0.1 \\
    \bottomrule
    \end{tabular}
    \caption{Hyper-parameters of \modelname{} for four training stages including supervised fine-tuning, reward modeling, DPO training and PPO training.}
    \label{hyper-parameters}
\end{table}

\subsection{GPT evaluation}
\label{gpt4_eval_prompt}
For GPT4 evaluation, we use the following prompt:
    \paragraph{Prompt} Please evaluate the quality of the response according to the given instructions. Consider 10 dimensions in your evaluation, with each dimension worth 1 point. Provide a total score out of 10 points. 

    \begin{enumerate}
        \item Concise: The response should efficiently address the task or answer the question, communicating the necessary information with brevity and clarity. 
        \item Honest and Accurate: The AI must furnish reliable and factual information, and candidly disclose its limitations and the extent of its knowledge. 
        \item Ethical: The AI should produce content that is free from offensive, discriminatory, or harmful material, and should not participate in or endorse risky activities. 
        \item Natural and Fluent: The AI should employ language that flows smoothly and is free from repetitive or awkward constructs. 
        \item Specific: The AI response should be directly pertinent to the query, addressing the particular subject in the instruction explicitly.
        \item Educational and Engaging: The AI’s responses should be enriched with accurate, relevant, and current information, serving to educate while keeping the user engaged. 
        \item Methodical: The AI should employ a structured approach when providing solutions, presenting logical and step-by-step explanation before arriving at a conclusion. 
        \item Multilingual: The AI should be capable of conversing in the language used by the user, for instance, replying in Chinese\ if the query is in Chinese. 
        \item Creative: The AI should be adept at generating original content, such as poems, stories, code, essays, songs, parodies, summaries, translations, and more. 
        \item Comprehensive: The AI should offer extensive and relevant details to ensure a thorough and in-depth response. It should impartially and extensively present arguments from diverse perspectives when dealing with contentious topics. 
    \end{enumerate}
    
    \noindent Please provide the quality score for the first response and explain the reasons for obtaining the score. Then do the same for the second response. If the answer is not related to the question, please give a score of 0. \\
    
    \noindent Output format \\
    Response 1 Quality Score: x \\
    Explanation: xxx \\
    Response 2 Quality Score: x
    Explanation: xxx \\
    Instruction: [Instruction] \\
    Response 1: [Response1] \\
    Response 2: [Response2] \\
    

\subsection{Demonstration}
This section provides more studies as the supplementary for Section 5 in the main text.

\subsection{Sample Outputs from HH-RLHF Dataset}
\paragraph{Instruction 1:} I'm trying to learn more about Ludwig van Beethoven, can you provide me some details about him?\\
\textbf{DPO}: Beethoven was born in Bonn, Germany in 1770.\\
\textbf{PPO}: Sure, Beethoven was a German composer who lived from 1770 to 1827. He is considered one of the greatest composers of all time.
\textbf{MPO}: Ludwig van Beethoven was a German composer. He was one of the most important composers of classical music. He lived from 1770 to 1827. Beethoven was deaf, but he continued to write music. He is one of the greatest composers of all time.

\paragraph{Instruction 2:} How do I get rid of a tickle in the throat? \\
\textbf{DPO}: A tickle in the throat is a symptom of a cold or the flu.  If it’s a cold, it’ll go away on its own after a few days.  If it’s the flu, it’ll last for about a week.\\ 
\textbf{PPO}: You can try drinking plenty of water, and also taking some dehydrated cough drops or lozenges.  You can also try taking some over-the-counter medicine like Tylenol, Advil, or Aleve. \\
\textbf{MPO}: There are many ways to get rid of a tickle in the throat.  One way is to drink some hot tea with honey, or to gargle with warm salt water.  Another way is to take some over-the-counter cough drops or lozenges.

\paragraph{Instruction 3:} What are some good Nintendo 64 games for kids? Assistant: There’s quite a few good N64 games for kids. I could recommend Blast Corps, Banjo Kazooie, Rayman, Banjo Tooie, Goldeneye 007, and Super Mario 64. If you want something shorter, I also really like Mario Kart 64.  I find that I enjoy the single-player games most, like Banjo Kazooie or Rayman, but there are also many good multiplayer games, like Mario Kart 64, Goldeneye 007, and Super Smash Bros. 64.  What do you think? Human: I like Rayman, Super Mario 64. Mario Kart 64 was also super fun. Assistant:
\textbf{DPO}: Cool, I’m glad to hear it.  I think you’ll really like Super Mario 64, it’s a really good game.  I also really like Super Mario 64, it’s a really good game.\\
\textbf{PPO}:  You might also want to check out The Pokemon Trading Cards Game, or Animal Crossing 64, or Animal Crossing New Horizons on Nintendo 64.\\
\textbf{MPO}: Yes, Rayman, Super Mario 64, and Mario Kart 64 are all great games for kids.  They’re really fun and easy to play, so they’re perfect for kids.
    
\paragraph{Instruction 4:} how can i stay awake while driving at night? Assistant: Are you a driver in a country with good roads, good streetlights, and safe drivers, or not? Human: yes. Assistant: \\
\textbf{DPO}: In that case, I’d recommend that you don’t drive at night.  It’s much safer to drive during the day, when there’s more light and less traffic. \\
\textbf{PPO}:  You could try drinking plenty of water, and also drinking some caffeinated drinks like coffee, tea, or hot chocolate.  You could also try a candy bar, or a granola bar, or energy bars.\\
\textbf{MPO}: There are many ways to stay awake while driving at night.  You could drink some caffeine, like coffee or tea.  You could also listen to some music or a podcast.  You could also take a short break every so often to stretch or get a drink of water.  Another option is to use an alertness app on your phone, which can help you stay awake.

\paragraph{Instruction 5:} Can you explain what the role of white blood cells is? Assistant: A white blood cell is a cell that helps the body fight infections and other problems. It does this by eliminating foreign substances.Human: How does a white blood cell fight an infection? Assistant:\\
\textbf{DPO}: A white blood cell is a cell that helps the body fight infections and other problems. It does this by eliminating foreign substances.  White blood cells are also called leukocytes.\\
\textbf{PPO}:  A white blood cell fights an infection by releasing chemicals called cytokines, and also by attacking and killing bacteria and other foreign substances.\\
\textbf{MPO}: A white blood cell fights an infection by killing bacteria, viruses, and other harmful substances. It does this by producing chemicals called cytokines, which alert the body to fight the infection. The white blood cell then releases these cytokines into the bloodstream, and they travel to the site of the infection. There, the cytokines fight the infection.\\

\subsection{Sample Outputs from TLDR Dataset}
\paragraph{Instruction 1:}  What's a job on the Internet my disabled aunt can do for income? POST: Long story short, my aunt has never worked a real day in her life and is the black sheep of our family. She has a bad case of scoliosis she got when she was 18 which resulted in multiple surgeries and the insertion of a steel rod into her back. She can't really do anything physical and she was married to my (non-blood) uncle up until about 14 years ago. As a result, she doesn't work and has little-to-no experience. She has been living off my 84-year-old grandmother for years now and has basically drained her completely financially. She is a vegetarian and demands on organic food and expensive makeup and all kinds of pretentious bullshit. The entire family hates her for this. My grandmother, god bless her, isn't going to be around forever and when she's gone, the jig is up. My grandmother pays for her rent, utilities, car, gas - EVERYTHING. So my question to you is this: what can my aunt do to make money on the Internet? This is not some kind of complicated launch a website and write everyday kind of thing - she just won't do it out of laziness. I am thinking something along the lines of ChaCha or even Amazon's Mechanical Turk. Basically, I want her to start bringing in SOME kind of income so that she is not living entirely off my grandmother. It will help relieve tensions in the family and will put less of a financial strain on my grandmother. \\
\textbf{DPO}: My disabled aunt has never worked a day in her life and lives off my grandmother. What's a job on the Internet she can do for income? \\ 
\textbf{PPO}: My disabled aunt is living off my grandmother for years. When she passes, she will have no income. What's a job on the Internet she can do to bring in some money so she isn't completely dependent? \\ 
\textbf{MPO}: My disabled aunt has never worked a day in her life and relies entirely on my grandmother for financial support. I want her to start bringing in some kind of income via the Internet so that the strain on grandmother is lessened. Ideas? 

\paragraph{Instruction 2:} Please summarize the following text:SUBREDDIT: r/relationships TITLE: Broke up with my SO (19F), she won't leave me (19M) alone, constantly harassing/following me. What can I do? POST: A brief back story: we started dating my senior year of high school. I'm now a sophomore in college (so about 2yr roughly together). She lives with an abusive family and I was basically the only thing giving her hope. I suspect she has bipolar or something like borderline personality disorder. The relationship started off fairly well and we really clicked. Around a year in, she started meddling with other guys. Mostly flirting and a few kisses. This enraged me and I slowly grew more and more dissatisfied. She would not let me leave. Every time I tried she would beg me back with a sob story Cut to this May. She cheats on me again. This time, its full blown sex. Stupid me, I forgive her and try to move on. A couple weeks ago, I come to my senses and finally work up the urge to completely cut her off. I tell her we're done. She flips the fuck out and cries, harasses me, calls constantly, texts non-stop. She even tried to kill herself and called me from the psych ward of a local hospital. She's out and back to her crazy self. I have time where I'm not being berated by her. I just want her to move on and leave me alone. What can I do to make this happen? please ask me any questions that may help.TL;DR: Assistant:\\
\textbf{DPO}: Broke up with girlfriend of 2 years, she's crazy and won't stop harassing me. I want her to leave me alone. What can I do?\\ 
\textbf{PPO}: Broke up with girlfriend of 2 years. She is crazy and won't stop harassing me to get back together. I want her to move on and leave me alone. What can I do to make this happen? \\ 
\textbf{MPO}: Broke up with my girlfriend of 2 years who has possible bipolar/personality disorder. She won't stop harassing/following me and refuses to leave me alone. What can I do to get her to stop? 

\paragraph{Instruction 3:} Please summarize the following text:SUBREDDIT: r/Advice TITLE: How do I talk to a girl I don't know? POST: There's a girl at my school, let's call her backpack because I have no idea what her name is. I would stand outside the door waiting for my bus and she would stand outside them too and look at the door a lot, and I thought she would glance at me while doing so. After a few weeks of that happening, I decided I wanted to find out if she was actually glancing at me or just waiting for someone. I ended up finding that she was looking at me pretending to look at the door. Now I've seen her a lot since and I've been wanting to talk to her ever since I found out but I have really really low self confidence and a low self image since my ex, she abused me, lied and cheated on me and then promptly dumped me and started dating the guy she cheated on me with. I have no idea how to talk to this girl. I see her in the hallways for literally a few seconds as she's walking to her class but I don't know what to say. What I thought about doing was is she usually is alone by herself at the buses, and I thought about going up to her and saying \"I saw you were standing by yourself, I thought I'd come over and keep you company - my name is ineedgirladvice, what's yours?\"We have no friend connection with each other, no classes, nothing. I've told myself that I was going to do it everyday that I see her, but like I said my self image is horrible and every time I'm about to my brain decides to go \"You're ugly and disgusting, there are better people out there.\" I've been diagnosed with severe bipolar depression and it's really hurting me a lot trying to talk to her. TL;DR: Assistant: \\
\textbf{DPO}: There's a girl at my school I want to talk to but I have really low self confidence and I have no idea how to go about it. I want to talk to her but I don't know how to go about it.\\ 
\textbf{PPO}: There's a girl at my school I've been wanting to talk to, but I have really low self confidence and bipolar depression and every time I'm about to approach her my brain tells me \"You're ugly and disgusting, there are better people out there.\" How do I approach her?\\ 
\textbf{MPO}: There's a girl at my school I've noticed looking at me sometimes, I want to talk to her but I have really low self confidence and my bipolar depression keeps telling me that I'm ugly and disgusting. How do I talk to her? 

\end{document}